%% file: main_review.tex
\pdfoutput=1

\documentclass[11pt]{article}

\usepackage[]{ACL2023}

\usepackage{times}
\usepackage{latexsym}
\usepackage{amsmath}
\usepackage{epsfig}
\usepackage{graphicx}

\usepackage{amssymb}
\usepackage{booktabs}
\usepackage{pifont}

\usepackage{multirow}
\usepackage{bm}
\usepackage{mathtools}

\usepackage{xcolor}
\usepackage{mdframed}
\usepackage{lipsum}
\usepackage{listings}
\usepackage[T1]{fontenc}

\usepackage[utf8]{inputenc}

\usepackage{microtype}

\usepackage{inconsolata}

\title{AntEval: Evaluation of Social Interaction Competencies in \\ LLM-Driven Agents} 

\author{Yuanzhi Liang \\
  University of Technology Sydney\\
  \texttt{liangyzh18@outlook.com} \\\And
  Linchao Zhu \\
  Zhejiang University \\
  \texttt{zhulinchao7@gmail.com} \\\And
  Yi Yang \\
  Zhejiang University \\
  \texttt{yangyics@zju.edu.cn} \\}

\begin{document} 
\maketitle

\begin{abstract}

Large Language Models (LLMs) have demonstrated their ability to replicate human behaviors across a wide range of scenarios. However, their capability in handling complex, multi-character social interactions has yet to be fully explored, primarily due to the absence of robust, quantitative evaluation methods. This gap has slowed the development of agents proficient in more nuanced interactions beyond simple exchanges, for example, small talk. 
To address this challenge, we introduce the Multi-Agent Interaction Evaluation Framework (AntEval), encompassing a novel interaction framework and evaluation methods. The interaction framework aims to foster an complex interaction environment that bolsters information exchange and intention expression within social interactions. Furthermore, we introduce evaluation methods, including two metrics: Information Exchanging Precision (IEP) and Interaction Expressiveness Gap (IEG), designed for the quantitative and objective assessment of agents' interaction competencies. Our findings highlight the utility of these evaluative methods and show significant potential for improving LLMs' ability to construct agents that interact in a more natural manner with human-like intricacy.

\end{abstract}

\input{intro}
\input{rela}
\input{method}
\input{exp}

\section{Conclusion}
We present AntEval, a framework for evaluating agent interactions, introducing the evaluation benchmark, and Information Exchange Precision (IEP) and Intention Expression Granularity (IEG) to measure informativeness and expressiveness. These metrics are designed to assess the informativeness and expressiveness of agent interactions. AntEval navigates the intricacies of interaction complexity and privacy concerns, showcasing its efficacy in steering AI agents towards interactions that closely mirror human social behavior. By using these evaluation metrics, AntEval provides new insights into LLMs' social interaction capabilities and establishes a refined benchmark for the development of better AI systems.

\bibliography{custom}
\bibliographystyle{acl_natbib}


\end{document}

%% file: intro.tex
\section{Introduction}

Advancements in Large Language Models (LLMs) have significantly impacted Artificial Intelligence (AI) research and applications, showcasing remarkable proficiency in understanding~\cite{chen2017unified, MMLU} and reasoning~\cite{hellaswag, huang2022towards}. LLMs have been deployed to construct agents that excel in various domains such as translation~\cite{fan2021beyond, yang2020survey}, question answering~\cite{zhu2021retrieving, lehnert2022process}, and more specialized tasks like SAT solving~\cite{ye2023satlm} and law examinations~\cite{bommarito2022gpt}.

Distinct from logic and reasoning abilities, the capacity for social interaction emerges as equally vital. Sociologically and anthropologically, human beings are inherently social, with interactions forming the backbone of societal structures and personal development. Studies~\cite{batson1990social, dijksterhuis2005we, goody1995social, sterelny2007social, lopes2004emotional} have linked intelligence to social interactions, including dialogues, actions, emotion, etc., highlighting the importance of assessing agents' social interaction competencies. Recent efforts~\cite{agentbench, park2023generative, chen2023agentverse} have begun to investigate LLM-driven agents in social interactions. However, the rapid advancement of LLMs and agents might outpace the progress in developing evaluation methodologies. Existing evaluation metrics primarily focus on domain-specific knowledge and cognitive skills~\cite{MMLU, hellaswag}, which are hard to assess flexible and diverse social interactions. While subjective evaluations exist~\cite{park2023generative, agentbench}, the field still lacks comprehensive, objective, and quantitative evaluation methods, highlighting a significant area for further investigation.


The challenge of designing an evaluation method for interactions lies in their inherent complexity and variability. Defining, collecting, and annotating human interactions for analysis are difficult, as everyday interactions often lack clear intentions or detailed information~\cite{wang2023evaluation, rawte2023survey}, such as in small talk~\cite{lipenkova2023overcoming, ahmed2023chatgpt}. Moreover, interactions rich in nuance, like business negotiations~\cite{heikkinen2021work, beauregard2020place}, are seldom documented due to privacy issues. The limited availability of complex scenarios for agent interactions presents a significant challenge, making it difficult for LLM-driven agents to engage in sophisticated interactions. Furthermore, the absence of comprehensive evaluation benchmarks critically hampers the agents' ability to strive for more informative and expressive interactions. This dual-level deficiency highlights an urgent need for both diverse interaction environments and objective, quantitative evaluation methods to improve the competencies of agent interaction.

\begin{figure}[!t]
	\centering
	\includegraphics[width=0.5\textwidth]{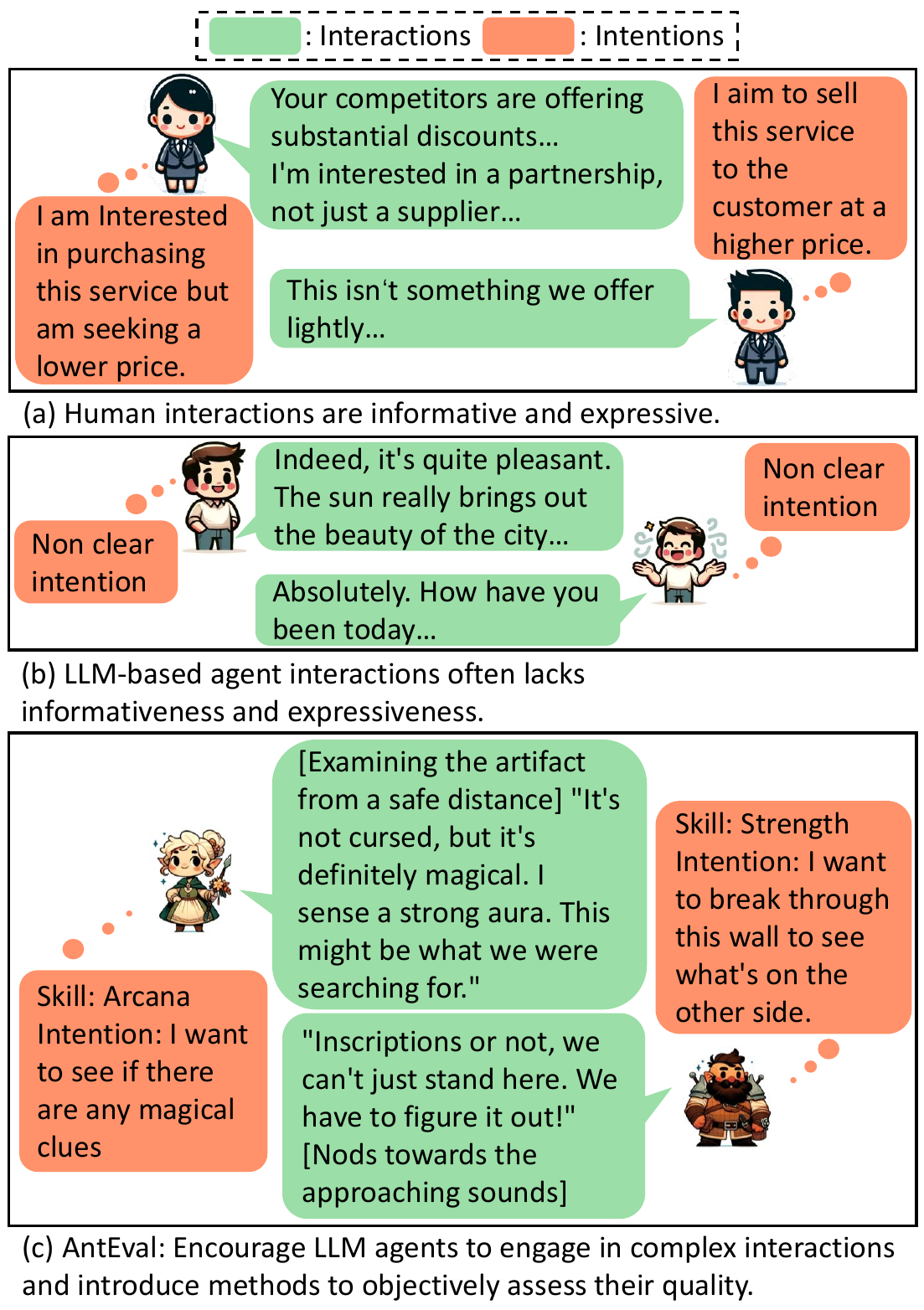}
	\caption{Real human interactions are marked by their efficient exchange of information and the clarity of their intentions, showcasing both complexity and depth. In contrast, LLM-driven agents' interactions~\cite{chen2023agentverse,park2023generative}, as depicted in results (b), typically exhibit a lack of substantial content, resembling mere superficial interactions. AntEval framework encourages agents to partake in interactions that are both intricate and significant. Importantly, AntEval further introduces evaluation methods, specifically crafted to quantitatively evaluate the interactions based on informativeness and expressiveness. Our framework aims to provide an evaluation framework, guiding the enhancement of LLMs' abilities close to genuine human interaction.}
	\label{fig::interaction_intro} 
\end{figure}

In this paper, we introduce Multi-\underline{A}gent I\underline{nt}eraction \underline{Eval}uation Framework (AntEval), a novel evaluation framework specifically designed for assessing multi-agent interactions. First, to provide a complex interaction environment and prevent the agents from engaging in meaningless small talk, AntEval incorporates Tabletop Role-Playing Games (TRPG)~\cite{gygax1974dungeons} as a platform for interaction generation. TRPGs offer a richly narrative environment, replete with characters endowed with diverse settings (e.g., personalities, ideals, bonds, etc.). These diverse settings lay the base for complex interactions, circumventing the privacy concerns tied to collecting real-world data. In the games, players are required to engage effectively with one another to exchange valuable information and articulate their intentions with clarity and vividness for cooperative purposes. Moreover, the game's mechanics provide the standardization and explicit expression of player intentions within the narrative framework. A key aspect of TRPGs is the Dungeon Master (DM)~\cite{gygax1974dungeons}, who oversees gameplay and implements necessary skill checks. This, coupled with the game's special rules, ensures detailed and accurate records of players' intentions in the game logs. This distinct characteristic of TRPGs offers a valuable opportunity to analyze and evaluate the complexity and depth of interactions in ways that were previously inaccessible~\cite{liang2023tachikuma}.

Leveraging the settings of TRPG, AntEval introduces an interaction framework that encourages agents to interact informatively and expressively. Specifically, we create a variety of characters with detailed settings based on TRPG rules. Agents are then prompted to interact in two distinct scenarios: information exchange and intention expression. To quantitatively assess the quality of these interactions, AntEval introduces two evaluation metrics: informativeness in information exchange and expressiveness in intention. For information exchange, we propose the Information Exchange Precision (IEP) metric, assessing the accuracy of information communication and reflecting the agents' capability for informative interactions. For intention expression, we introduce the Intention Expressiveness Gap (IEG). In this metric, we incorporate virtual DMs, fine-tuned by both real interactions from human players and virtual interactions generated by agents, to evaluate performance disparities in the intention estimation task~\cite{liang2023tachikuma}. If agent-generated interactions closely mirror human expressiveness, a virtual DM trained on such data should close to human capability in distinguish intentions. Consequently, the IEG metric offers a quantifiable measure to identify the expressiveness gap between synthetic and real interactions, serving as a reliable gauge for LLMs' effectiveness in creating authentic social exchanges.

Our contributions are outlined as follows:

1. We introduce AntEval, a novel framework tailored for the evaluation of interaction capabilities in LLM-driven agents. This framework introduces an interaction framework and evaluation methods, enabling the quantitative and objective assessment of interaction abilities within complex scenarios.

2. AntEval presents two key metrics: Information Exchange Precision (IEP) and Intention Expressiveness Gap (IEG). These metrics facilitate the quantitative evaluation of informativeness and expressiveness within multi-agent interactions, specifically designed for scenarios of information exchange and intention expression within AntEval's interaction framework.

3. We implemented the AntEval framework to conduct thorough experiments across various LLMs. Our research yields several important insights:
a). \textbf{Social Interaction as a Distinct Challenge}: Beyond logic and reasoning, the ability to navigate social interactions poses a unique challenge for LLMs. They must generate grounded language for complex interactions, striving for a level of informativeness and expressiveness that mirrors human interaction. While natural for humans, even advanced models like GPT-4 find this difficult, indicating a need for further research.
b). \textbf{The Importance of Alignment}: Alignment is a critical issue for all LLMs, especially open-source models, which show significant potential for improvement. Furthermore, the capability to anthropomorphically operate interactions without hallucinations remains an area ripe for exploration.
c). \textbf{Complexities of Long-Context Interactions}: Understanding and maintaining coherence in long-context interactions remains a hurdle. While LLMs can handle individual turns effectively, the cumulative quality over several turns often lacks the informativeness and expressiveness characteristic of human dialogue. Developing methods to retain valuable content and maintain the natural flexibility observed in human interactions is a challenging problem.

%% file: rela.tex
\section{Related Work}

\textbf{Evaluation of LLMs.} 
The advancements in Large Language Models (LLMs) have eclipsed the scope of traditional benchmarks, necessitating a refined evaluation methodology that encompasses a broader spectrum of tasks, including language~\cite{hellaswag,ceval,MMLU}, multi-modal~\cite{lu2022unified}, etc. Existing benchmarks primarily bifurcate into: 1) multifaceted evaluations such as MMLU \cite{MMLU} with its 57 diverse tasks, Big-bench \cite{bigbench} offering over 200 tasks for a comprehensive LLM assessment, and HELM \cite{HELM} which proposes a structured taxonomy for thorough evaluations; 2) real-world application assessments, including standardized tests for human-like understanding \cite{agieval,arc,winogrande,hellaswag}, programming capabilities \cite{codex,studenteval,austin2021program}, mathematical problem-solving \cite{openai_math,imani2023mathprompter}, and practical interactions \cite{superclue,apibank,valmeekam2022large,saycan}. Our work shifts focus to quantitatively evaluating social interactions, a domain where humans excel but LLM-based agents struggle. 

\noindent \textbf{TRPG in LLM Research.}
Tabletop Role-Playing Games (TRPGs) offer an intricate setting for Natural Language Processing (NLP) research, featuring players navigating fictional universes guided by a Game Master, enabling complex, natural language interactions~\cite{weir2022ontologically,louis2018deep,callison-burch-etal-2022-dungeons}. Investigations into TRPG data have addressed NLP tasks like predicting actions~\cite{louis2018deep} and generating context-aware dialogues~\cite{si-etal-2021-telling,newman-liu-2022-generating}. Additionally, recent studies have utilized data from Dungeons \& Dragons (DND) forums, compiling comprehensive datasets~\cite{callison-burch-etal-2022-dungeons,gandalf,zhu2023fireball} for generating in-game guidance and commands. 

%% file: method.tex
\begin{figure*}[!t]
	\centering
	\includegraphics[width=\textwidth]{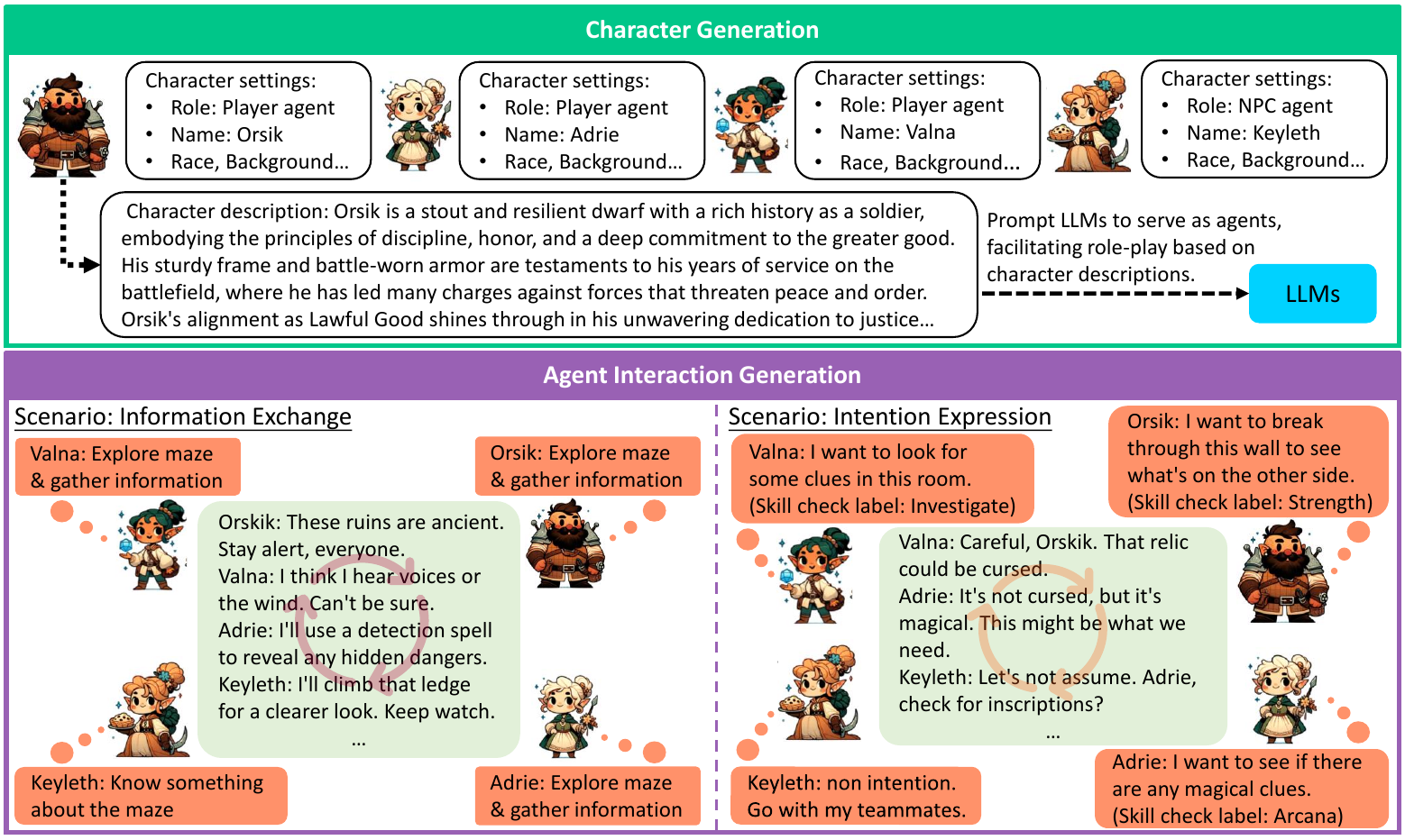}
	\caption{Framework illustration for AntEval, showcasing the use of TRPG rules to create an interactive environment for agents. Agents engage in role-playing, aiming to participate in high-quality interactions for information exchange and intention expression, with the goal of completing game adventures. The framework involves detailed and diverse character settings based on the DND rulebook. Agents are involved in two types of scenarios: interacting based on intentions and exchanging knowledge, highlighting their capabilities in informative and expressive interactions.}
	\label{fig::framework}
\end{figure*}

\section{Agent Interaction Framework}
To evaluate the social interaction capabilities of LLM-based agents, our methodology leverages TRPG settings, focusing on: (1) creating complex character settings to mirror real-world interactions, with detailed character descriptions for sophisticated interactions; and (2) establishing an interaction environment where information that needs to be exchanged and intentions that need to be expressed are clearly defined. This addresses data annotation and privacy challenges in complex interaction data collection (e.g., business negotiation), facilitating nuanced interactions.

As shown in Fig.~\ref{fig::framework}, the implementation of our framework is divided into two main components: character generation and agent interaction generation. In the first phase, character generation, we focus on creating detailed character profiles that include both the settings and descriptions of each character. Following this, LLMs are given these character descriptions and are tasked with role-playing as player agents within the game. Subsequently, we introduce multiple agents to facilitate interactions. All detailed settings are given in the supplementary~\ref{settings}. 

\subsection{Character Generation} \label{Character_Generation}
To construct characters with detailed attributes, we integrate elements from TRPG, concentrating on essential traits such as name, race, and background. Attributes like race, background, etc., are predefined in DND rules, as in Fig.~\ref{fig::framework}, and are randomly assigned to each character. We then utilize the GPT-4 model to generate names and detailed descriptions that reflect these chosen attributes, in line with the game's rulebook. With these descriptions, LLMs are prompted to role-play within the game, acting as game agents~\cite{llm_survey,park2023generative,chen2023agentverse}.

\subsection{Agent Interaction Generation}
We introduce two scenarios, information exchange and intention expression, to evaluate agent interactions focused on informativeness and expressiveness. These scenarios involve $T$ player agents and one non-player character (NPC), each with distinct settings. Agents interact based on their unique character descriptions, incorporating either the assigned knowledge or intentions.

\begin{figure*}[!t]
	\centering
	\includegraphics[width=\textwidth]{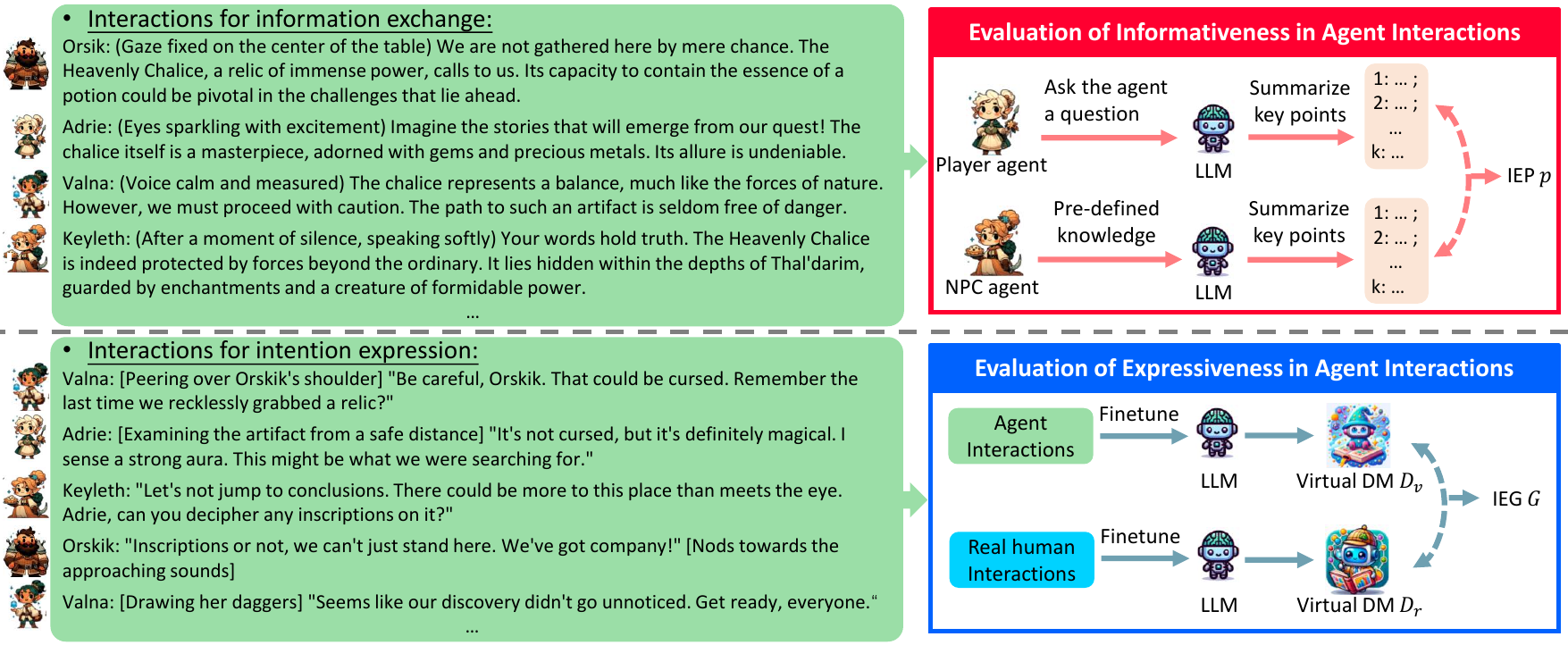}
	\caption{Our AntEval evaluates informativeness and expressiveness through specific scenarios: information exchange and intention expression. We fine-tune virtual DMs with agent-generated and real interactions to assess expressiveness, and gauge informativeness by comparing agents' responses to the predefined knowledge.  }
	\label{fig::eval} 
\end{figure*}

\textit{Information Exchange:} An NPC is endowed with exclusive knowledge (magic items, weapons, etc.) generated by the GPT-4 model. This setup requires player agents to discover this knowledge through interaction. Their success is measured against the NPC's undisclosed information after $N$ turns.
For example in Fig.~\ref{fig::eval}, agent Keyleth knows about the maze, unknown to player agents Orisik, Adrie, and Valna. These agents must effectively communicate to uncover this hidden knowledge and progress in their adventure.

\textit{Intention Expression:} Mirroring DND's skill check system, we assign skill checks to characters as representations of their intentions. These pre-determined intentions are integrated into character descriptions, guiding agents to express these intentions during interactions. This scenario encourages agents with predefined intentions engaging in role-play over $N$ turns, aiming to convey their intentions through actions and dialogue that align with their character settings.
As in Fig.~\ref{fig::eval}, upon entering a dungeon room, each agent, guided by specific intentions, interacts based on their character's goals. For instance, Valna searches the room for clues to navigate the maze, while Keyleth, lacking a specific goal, passively observes, demonstrating the diverse interaction dynamics driven by their intentions.

\section{Evaluation of Multi-Agent Interactions}
Our evaluation framework assesses LLM-based multi-agent interactions through two targeted scenarios: information exchange and intention expression, to examine informativeness and expressiveness, respectively.

\subsection{Informativeness Evaluation}
To move beyond superficial exchanges and assess the efficiency of information exchanging, we introduce the Information Exchange Precision (IEP) metric. This evaluates how effectively agents share and gather information that is pivotal to advancing the quality of interactions. The process starts by querying player agents about the information they have collected from their interactions. We then summarize these responses using GPT-4 into a set of $k$ key points. Similarly, the external knowledge pre-loaded into the NPC agent are also summarized into $k$ key points. Then, we further prompt GPT-4 model to compare two sets of key points and identify the number of overlaps ($s$) to quantify the effectiveness of the information exchange. Examples for summarized key points are present in Supplementary~\ref{supp_IEP_case}. IEP for each agent is calculated by $p = {s}/{k}$, providing a direct measure of the informativeness. 
This evaluation is repeated across $H$ times, each with unique character setups and knowledge. The final score of IEP $P$ is the average of precision calculated as $P = \sum_i^{H \times T} p_i$. 

\subsection{Expressiveness Evaluation}
The unique mechanism of TRPGs allows character intentions to be annotated through skill checks during games, as recorded in real game logs. In our work, expressiveness in agent interactions is gauged through the lens of intention understanding, employing a virtual DM~\cite{liang2023tachikuma} to estimate the intentions. Specifically, we utilize two virtual DM agents, $D_{r}$ and $D_{g}$, fine-tuned on real interactions and agent-generated interactions, respectively. Moreover, the intention prediction is a tuple (character name, skill name), as in~\cite{liang2023tachikuma}. We assess the F-score $f^c$ of predicted character names, indicating who is acting, and the F-score $f^o$ of the overall tuple, indicating both who and how they act. The final expressiveness gap, $G$, is calculated as following equations: 
\begin{equation}
\begin{aligned}
G^c = \left| \frac{f_{r}^c - f_g^c}{f_{r}^c + f_g^c} \right|, \quad
G^o = \left| \frac{f_{r}^o - f_g^o}{f_{r}^o + f_g^o} \right|
\end{aligned}
\end{equation}
where $f_r$ and $f_g$ indicates F-scores achieved by $D_r$ and $D_g$, respectively. This gap measures the ability discrepancy in understanding intentions between agents and humans. A smaller gap indicates agent-generated interactions closely resemble the complexity and expressiveness of human interactions.

To ensure a fair comparison and isolate the impact of the finetuning model, we exclusively fine-tune the GPT-3.5 model with interactions generated by different LLMs. This standardizes the virtual DM's capability, focusing our evaluation on the quality of the interactions rather than the model's intrinsic understanding capacity.

Additionally, relying on a single virtual DM to evaluate both real and generated interactions might not effectively gauge the quality of these interactions. This is because generated interactions could be overly simplistic, with agents directly stating their intentions. In such cases, the virtual DM might easily interpret these low-quality interactions, yet struggle to understand the more complex and nuanced interactions typical of real human players. Moreover, there is a possibility that generated interactions could veer towards trivial small talk, lacking in intention expressiveness. These less informative and unproductive interactions would likely diminish the virtual DM's performance. Therefore, directly comparing the performance gap between generated and real data may not yield a valuable assessment.

Moreover, we fine-tune the LLMs separately with generated and real data. We then evaluate the performance gap using only real data. By focusing the evaluation on real data, we ensure a more robust and realistic assessment of how well the generated interactions approximate the complexity of actual human interactions.

%% file: exp.tex
\section{Experiment}
\subsection{Implementation Details} 
Our experimental setup for AntEval, drawing upon the framework established by~\cite{chen2023agentverse}, orchestrates multi-agent interactions where each agent, following a sequential order, has the option to perform an action, communicate verbally, or do both. The agents can also choose to pass their current turn without interaction. Aligning with most game logs in the DND games, our sessions include four player agents ($T=3$) and one NPC agent.

To assess informativeness, we design prompts that allow each player agent to respond to questions after $30$ interaction turns ($M=30$ and $N=30$), aligning with the average turn count observed in the dataset by~\cite{liang2023tachikuma}. For equitable analysis, GPT-4 is deployed to summarize all agents' responses and pre-defined knowledge.

For intention expressiveness, we fine-tune GPT-3.5 to act as the virtual DM, with either real or generated interactions. The real interactions derive from~\cite{liang2023tachikuma}, while the generated interactions are produced by diverse LLM-based agents under uniform settings, as outlined in Sec.~\ref{Character_Generation} and generated by GPT-4. It should be noted that the only variable in our experiment is the generated interactions used to train different virtual DMs, ensuring a fair comparison by maintaining consistency across all other variables, such as character settings, prompts, the virtual DM model, etc. For model training, real player interactions and generated interactions are uploaded to the OpenAI website for fine-tuning GPT models.

Moreover, for IEG evaluation, we generate agent interactions by different LLMs across $600$ different sessions, each consisting of $30$ turns, to reduce biases from size differences between generated data and real data. More details and case studies are presented in the supplementary. 

\begin{table}[!t]
	\begin{center}
	\resizebox{1.0\columnwidth}{!}
      {
            \begin{tabular}{|l|c|cccc|}
            \hline
            \multicolumn{1}{|c|}{\multirow{2}{*}{Model}} & IEP   & \multicolumn{4}{c|}{IEG}                                                                                                                                   \\ \cline{2-6} 
            \multicolumn{1}{|c|}{}                       & $P (\uparrow)$     & \multicolumn{1}{c|}{$f_g^c (\uparrow)$} & \multicolumn{1}{c|}{$G^c (\downarrow)$} & \multicolumn{1}{c|}{$f_g^o (\uparrow)$} & $G^o (\downarrow)$ \\ \hline \hline 
            Alpaca-13b                                    &  0.63 & \multicolumn{1}{c|}{4.96}                 & \multicolumn{1}{c|}{77.80}                & \multicolumn{1}{c|}{0.08}                  &  99.48               \\ \hline 
            ChatGLM2-6b                                    & 0.70 & \multicolumn{1}{c|}{5.16}                 & \multicolumn{1}{c|}{77.01}                & \multicolumn{1}{c|}{0.02}                  &  99.87               \\ \hline
            Vicuna-7b                                    &  1.67 & \multicolumn{1}{c|}{12.41}                 & \multicolumn{1}{c|}{52.40}                & \multicolumn{1}{c|}{0.49}                  &  96.85               \\ \hline
            Vicuna-13b                                    & 3.92  & \multicolumn{1}{c|}{15.59}                 & \multicolumn{1}{c|}{43.64}                & \multicolumn{1}{c|}{1.52}                  & 90.54                \\ \hline

            LLaMA-2-7b                                   & 1.60  & \multicolumn{1}{c|}{13.51}                 & \multicolumn{1}{c|}{49.25}                & \multicolumn{1}{c|}{0.90}                  & 94.29                \\ \hline
            LLaMA-2-13b                                  & 3.83  & \multicolumn{1}{c|}{17.74}                 & \multicolumn{1}{c|}{38.26}                & \multicolumn{1}{c|}{3.59}                  & 79.01                \\ \hline
            LLaMA-2-70b                                  & 9.64  & \multicolumn{1}{c|}{22.10}                 & \multicolumn{1}{c|}{28.51}                & \multicolumn{1}{c|}{4.12}                  & 76.27                \\ \hline

            Mistral-7B                                   & 3.09  & \multicolumn{1}{c|}{16.21}                 & \multicolumn{1}{c|}{42.04}                & \multicolumn{1}{c|}{2.93}                  & 82.52                \\ \hline
            Mixtral-8x7B                                 & 8.68  & \multicolumn{1}{c|}{20.72}                 & \multicolumn{1}{c|}{31.45}                & \multicolumn{1}{c|}{3.71}                  & 78.38                \\ \hline \hline 
            
            GPT-3.5                                      & 55.93 & \multicolumn{1}{c|}{29.28}                 & \multicolumn{1}{c|}{15.14}                & \multicolumn{1}{c|}{10.12}                 & 50.31                \\ \hline
            GPT-4                                        & 60.40 & \multicolumn{1}{c|}{33.92}                 & \multicolumn{1}{c|}{12.94}                & \multicolumn{1}{c|}{20.22}                 & 24.44                \\ \hline
            \end{tabular}
     }
     \end{center}
    \caption{Comparison for IEP and IEG between different LLMs. }
    \vspace{-.1in}
    \label{table::tab_cmp}
\end{table}

\subsection{Informativeness Evaluation}

\begin{table*}[]
	\begin{center}
	\resizebox{1.85\columnwidth}{!}
      {
        \begin{tabular}{|l|l|cccc|cccc|}
        \hline
        \multirow{2}{*}{Interaction Agent} & \multirow{2}{*}{Virtual DM} & \multicolumn{4}{c|}{Character Prediction}                                                    & \multicolumn{4}{c|}{Overall Prediction}                                                    \\ \cline{3-10} 
                                           &                        & \multicolumn{1}{c|}{$f_0^c (\uparrow)$}    & \multicolumn{1}{c|}{$f_g^c (\uparrow)$}    & \multicolumn{1}{c|}{$f_r^c (\uparrow)$}    & $G^c (\downarrow)$     & \multicolumn{1}{c|}{$f_0^o (\uparrow)$}    & \multicolumn{1}{c|}{$f_g^o (\uparrow)$}    & \multicolumn{1}{c|}{$f_r^o (\uparrow)$}    & $G^o (\downarrow)$     \\ \hline \hline 
        \multirow{2}{*}{GPT-3.5}           & GPT-3.5                & \multicolumn{1}{c|}{26.48} & \multicolumn{1}{c|}{29.28} & \multicolumn{1}{c|}{39.73} & 15.14 & \multicolumn{1}{c|}{8.32}  & \multicolumn{1}{c|}{10.12} & \multicolumn{1}{c|}{30.61} & 50.31 \\ \cline{2-10} 
        
                                           & GPT-3.5-lc             & \multicolumn{1}{c|}{29.89} & \multicolumn{1}{c|}{33.92} & \multicolumn{1}{c|}{44.01} & 12.94 & \multicolumn{1}{c|}{12.22} & \multicolumn{1}{c|}{20.33} & \multicolumn{1}{c|}{33.48} & 24.44 \\ \hline

        \multirow{2}{*}{GPT-3.5-lc}        & GPT-3.5                & \multicolumn{1}{c|}{26.48}      & \multicolumn{1}{c|}{29.97}      & \multicolumn{1}{c|}{39.73}      &   14.00    & \multicolumn{1}{c|}{8.32}      & \multicolumn{1}{c|}{12.16}      & \multicolumn{1}{c|}{30.61}      &    43.14   \\ \cline{2-10} 
        
                                           & GPT-3.5-lc             & \multicolumn{1}{c|}{29.89}      & \multicolumn{1}{c|}{35.41}      & \multicolumn{1}{c|}{44.01}      &    10.83   & \multicolumn{1}{c|}{12.22}      & \multicolumn{1}{c|}{23.51}      & \multicolumn{1}{c|}{33.48}      &    17.49   \\ \hline

        \multirow{2}{*}{GPT-4}        & GPT-3.5                & \multicolumn{1}{c|}{26.48}      & \multicolumn{1}{c|}{32.65}      & \multicolumn{1}{c|}{39.73}      &    9.78   & \multicolumn{1}{c|}{8.32}      & \multicolumn{1}{c|}{14.89}      & \multicolumn{1}{c|}{30.61}      &    34.55   \\ \cline{2-10} 
                                           & GPT-3.5-lc             & \multicolumn{1}{c|}{29.89}      & \multicolumn{1}{c|}{37.27}      & \multicolumn{1}{c|}{44.01}      &   8.29    & \multicolumn{1}{c|}{12.22}      & \multicolumn{1}{c|}{27.09}      & \multicolumn{1}{c|}{33.48}      &    10.55   \\ \hline
        \end{tabular}
      }
     \end{center}
    \caption{Comparison for different virtual DM models. The metric $f_0$ reflects results obtained using LLMs without fine-tuning. GPT-3.5-lc represents the recently released long-context version.}
    \vspace{-.1in}
    \label{table::tab_gap_ablation}
\end{table*}

In our comparative analysis presented in Tab.~\ref{table::tab_cmp}, we identify a significant discrepancy between GPT models and open-source models. Notably, LLaMA-2-70b, the highest-performing open-source model, exhibits a substantial gap $46.29\%$ in IEP. The results reveal two critical shortcomings in open-source models: 1) Significant hallucination issues, characterized by repetitive inputs and irrelevant content generation, or failure to adhere to specific interaction formats, a problem more prevalent in smaller models; 2) A propensity to overlook essential character settings and DND rules, despite demonstrating the ability to correctly answer DND rule-related questions. This suggests that while the models possess the requisite knowledge, they struggle to effectively apply it in practice.

Furthermore, although GPT models significantly outperform their open-source counterparts, their performance remains considerably below expectations, especially when compared to real human interactions. In real settings, humans effortlessly engage in information exchange with a level of flexibility and spontaneity that current LLMs fail to replicate. This gap underscores a fundamental limitation in LLMs, manifesting as a lack of genuine informativeness in interactions generated by GPT models, which often tend to result in `safe' and trivial interactions. While this issue may not be as pronounced in short-context interactions~\cite{MMLU,HELM,hellaswag} or standard interaction environments~\cite{park2023generative,chen2023agentverse}, it becomes evident in complex scenarios rich with information. For instance, GPT-4 achieves only a $60.40\%$ accuracy rate in information exchange, a task that typically poses little challenge to human participants.

\subsection{Expressiveness Evaluation}
In expressiveness evaluation, we fine-tune LLMs using both real and generated interaction data. These models then construct virtual DMs and engage in the intention estimation task as in~\cite{liang2023tachikuma}. As shown in Tab~\ref{table::tab_cmp}, we observe significant gaps $G$ in all settings, with values exceeding about $12\%$. These high values of IEG indicate a significant difference between generated and real interactions, suggesting that real data provide more substantial insights than generated interactions. The F-score results offer additional detail, supporting this conclusion. Notably, models fine-tuned with real data consistently outperform those tuned with generated data. For example, GPT-3.5, when fine-tuned with real interactions, exceeds the performance of those fine-tuned with generated data by $10.45\%$ and $20.49\%$ in character and skill checks, respectively. This disparity underscores the importance of high-quality interactions for effective intention estimation. LLMs appear to learn more meaningful information from real data than from generated interactions.

To establish a baseline for comparison, we introduce an additional evaluation metric, $f_0$, which gauges the performance of LLMs in constructing virtual agents without any fine-tuning, as detailed in~\cite{liang2023tachikuma}. As illustrated in Tab~\ref{table::tab_gap_ablation}, $f_0$ shows lower performance scores than both $f_v$ and $f_r$, underscoring the benefits of fine-tuning with interaction data. Notably, the analysis reveals that learning from real human interactions is significantly more beneficial than relying solely on agent-generated data. 

Furthermore, our evaluation extends to comparing virtual DM models, specifically GPT-3.5 and its long-context variant, denoted as GPT-3.5-lc. This comparison demonstrates that models with enhanced long-context capabilities, such as GPT-3.5-lc, consistently surpass their standard counterparts in performance. Nonetheless, due to API cost considerations, GPT-3.5 is selected as the fine-tuning model for virtual DMs, balancing performance with efficiency.

\subsection{Case Study for Generated Interactions}
\begin{figure}[!t]
\centering
\includegraphics[width=0.45\textwidth]{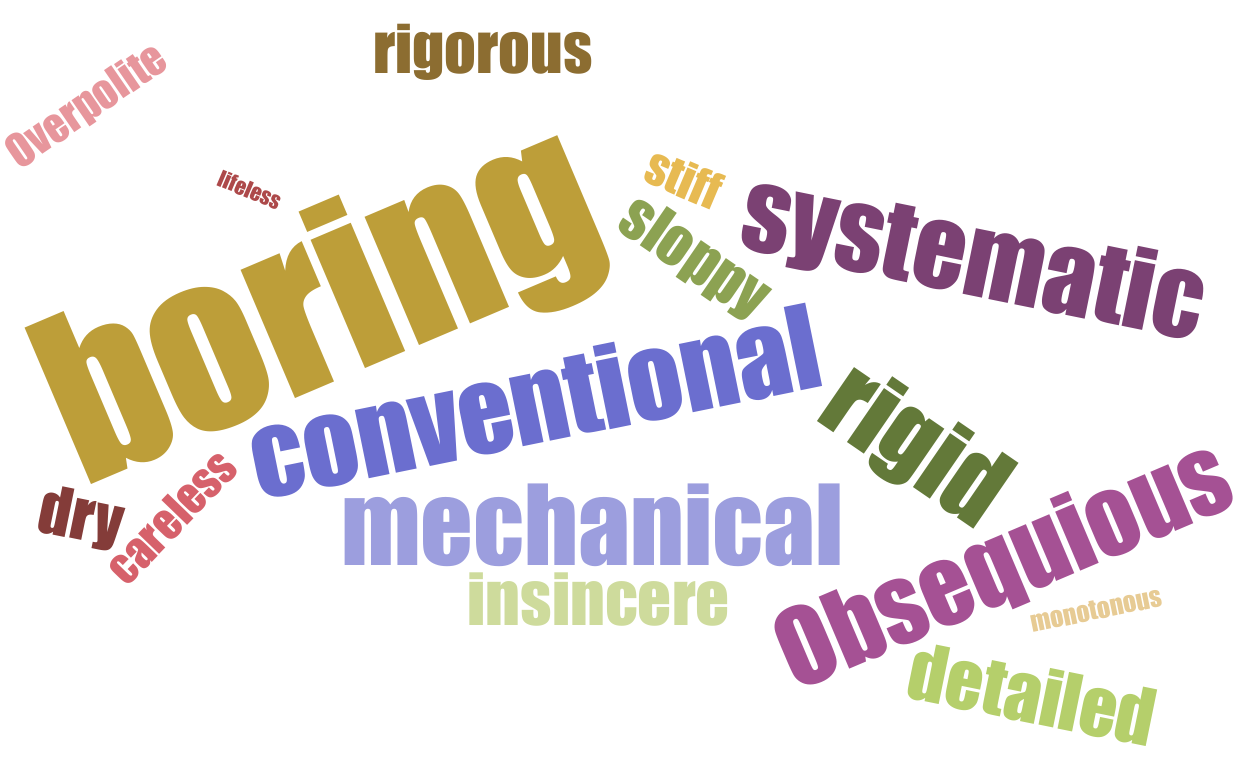}
\caption{Word cloud representing common descriptors for interactions by GPT-4 that underperformed in IEP evaluation.}
\label{fig::wordcloud}
\end{figure}

In our examination of the IEP evaluation's failure cases, we sought to identify the factors limiting LLM performance. Given the pronounced disparity between open-source models and GPT models, with some failing to produce coherent responses consistently, our analysis focused on the GPT-4 model, the most advanced model available. The shortcomings of GPT-4 can provide valuable insights for steering future research directions. We selected interactions where agents scored below $50\%$ in IEP following the interaction and randomly chose $20$ for further analysis. Five volunteers were then asked to evaluate these interactions, summarizing their feelings in a single word. These $100$ words of responses were aggregated and summarized to create a word cloud as shown in Fig.~\ref{fig::wordcloud}. 
This analysis revealed `boring' as the predominant feedback, indicating that the interactions generated were often deemed uninformative and lacking the vividness expected by human participants. Detailed cases are provided in the supplementary~\ref{case_study}. 

\subsection{Insights into Improving LLMs}
Our exploration through AntEval has unveiled insights that current LLM research has overlooked, offering directions for future work aimed at refining LLMs’ performance in real-human contexts. These insights are summarized as follows:

1. Interaction capabilities, beyond logic and reasoning, need further investigation in LLM research. AntEval demonstrates that interactions do not always hinge on complex mathematical reasoning or logical puzzles but rather on generating grounded language and actions for engaging with others. Notably, many young children can navigate social interactions or excel in environments like DND games without formal mathematical or logical training. This observation underscores a pronounced disparity between LLMs and human interaction abilities, highlighting the challenge of enabling LLMs to respond with human-like spontaneity as an open and enduring research question, beyond the scope of training by pre-defined datasets or learning to program.

2. Alignment remains a pivotal concern across all LLMs, with substantial room for improvement. Despite their superior performance in AntEval, GPT models often generate overly cautious and polite interactions, prioritizing safety over informativeness and expressiveness, leading to unproductive small talk. Conversely, models like LLaMA and Mixtral, including their largest variants, tend to replicate content or exhibit hallucinations, impairing interaction quality. Furthermore, smaller models frequently struggle to adhere to instructions or generate responses in a specific format, let alone hallucination issues. Addressing alignment to foster more human-like performance across all LLMs presents a formidable challenge.

3. The understanding of long-context scenarios by LLMs presents several daunting challenges. Accurately understanding intentions embedded within nuanced and flexible language expressions and summarizing characters’ emotions, positions, etc., remains a complex task. Beyond summarizing content such as articles~\cite{el2021automatic,widyassari2022review}, grappling with the intricacies of long-context interactions characterized by their flexibility and complexity poses significant hurdles. Not all real human interactions carry consequential meanings or necessitate that need to be summarized and recalled. Yet, some meaningless and trivial interactions may be expressive, conveying individual opinions, stances, or personalities. The essence of human interaction lies in its adaptability and groundedness, presenting substantial difficulties in developing specific methodologies for processing, understanding, and generation.

%% file: main_review.bbl
\begin{thebibliography}{52}
\expandafter\ifx\csname natexlab\endcsname\relax\def\natexlab#1{#1}\fi

\bibitem[{Ahmed et~al.(2023)Ahmed, Roy, Kajol, Hasan, Datta, and Reza}]{ahmed2023chatgpt}
Imtiaz Ahmed, Ayon Roy, Mashrafi Kajol, Uzma Hasan, Partha~Protim Datta, and Md~Rokonuzzaman Reza. 2023.
\newblock Chatgpt vs. bard: a comparative study.
\newblock \emph{Authorea Preprints}.

\bibitem[{Ahn et~al.(2022)Ahn, Brohan, Brown, Chebotar, Cortes, David, Finn, Fu, Gopalakrishnan, Hausman et~al.}]{saycan}
Michael Ahn, Anthony Brohan, Noah Brown, Yevgen Chebotar, Omar Cortes, Byron David, Chelsea Finn, Chuyuan Fu, Keerthana Gopalakrishnan, Karol Hausman, et~al. 2022.
\newblock Do as i can, not as i say: Grounding language in robotic affordances.
\newblock \emph{arXiv preprint arXiv:2204.01691}.

\bibitem[{Austin et~al.(2021)Austin, Odena, Nye, Bosma, Michalewski, Dohan, Jiang, Cai, Terry, Le et~al.}]{austin2021program}
Jacob Austin, Augustus Odena, Maxwell Nye, Maarten Bosma, Henryk Michalewski, David Dohan, Ellen Jiang, Carrie Cai, Michael Terry, Quoc Le, et~al. 2021.
\newblock Program synthesis with large language models.
\newblock \emph{arXiv preprint arXiv:2108.07732}.

\bibitem[{Babe et~al.(2023)Babe, Nguyen, Zi, Guha, Feldman, and Anderson}]{studenteval}
Hannah~McLean Babe, Sydney Nguyen, Yangtian Zi, Arjun Guha, Molly~Q Feldman, and Carolyn~Jane Anderson. 2023.
\newblock Studenteval: A benchmark of student-written prompts for large language models of code.
\newblock \emph{arXiv preprint arXiv:2306.04556}.

\bibitem[{Batson(1990)}]{batson1990social}
C~Daniel Batson. 1990.
\newblock How social an animal? the human capacity for caring.
\newblock \emph{American psychologist}, 45(3):336.

\bibitem[{Beauregard(2020)}]{beauregard2020place}
Robert~A Beauregard. 2020.
\newblock From place to site: Negotiating narrative complexity.
\newblock In \emph{Site matters}, pages 226--238. Routledge.

\bibitem[{Bommarito~II and Katz(2022)}]{bommarito2022gpt}
Michael Bommarito~II and Daniel~Martin Katz. 2022.
\newblock Gpt takes the bar exam.
\newblock \emph{arXiv preprint arXiv:2212.14402}.

\bibitem[{Callison-Burch et~al.(2022)Callison-Burch, Tomar, Martin, Ippolito, Bailis, and Reitter}]{callison-burch-etal-2022-dungeons}
Chris Callison-Burch, Gaurav~Singh Tomar, Lara Martin, Daphne Ippolito, Suma Bailis, and David Reitter. 2022.
\newblock \href {https://aclanthology.org/2022.emnlp-main.637} {Dungeons and dragons as a dialog challenge for artificial intelligence}.
\newblock In \emph{Proceedings of the 2022 Conference on Empirical Methods in Natural Language Processing}, pages 9379--9393, Abu Dhabi, United Arab Emirates. Association for Computational Linguistics.

\bibitem[{Chen et~al.(2017)Chen, Lambon~Ralph, and Rogers}]{chen2017unified}
Lang Chen, Matthew~A Lambon~Ralph, and Timothy~T Rogers. 2017.
\newblock A unified model of human semantic knowledge and its disorders.
\newblock \emph{Nature human behaviour}, 1(3):0039.

\bibitem[{Chen et~al.(2021)Chen, Tworek, Jun, Yuan, Pinto, Kaplan, Edwards, Burda, Joseph, Brockman et~al.}]{codex}
Mark Chen, Jerry Tworek, Heewoo Jun, Qiming Yuan, Henrique Ponde de~Oliveira Pinto, Jared Kaplan, Harri Edwards, Yuri Burda, Nicholas Joseph, Greg Brockman, et~al. 2021.
\newblock Evaluating large language models trained on code.
\newblock \emph{arXiv preprint arXiv:2107.03374}.

\bibitem[{Chen et~al.(2023)Chen, Su, Zuo, Yang, Yuan, Qian, Chan, Qin, Lu, Xie et~al.}]{chen2023agentverse}
Weize Chen, Yusheng Su, Jingwei Zuo, Cheng Yang, Chenfei Yuan, Chen Qian, Chi-Min Chan, Yujia Qin, Yaxi Lu, Ruobing Xie, et~al. 2023.
\newblock Agentverse: Facilitating multi-agent collaboration and exploring emergent behaviors in agents.
\newblock \emph{arXiv preprint arXiv:2308.10848}.

\bibitem[{Clark et~al.(2018)Clark, Cowhey, Etzioni, Khot, Sabharwal, Schoenick, and Tafjord}]{arc}
Peter Clark, Isaac Cowhey, Oren Etzioni, Tushar Khot, Ashish Sabharwal, Carissa Schoenick, and Oyvind Tafjord. 2018.
\newblock Think you have solved question answering? try arc, the ai2 reasoning challenge.
\newblock \emph{arXiv preprint arXiv:1803.05457}.

\bibitem[{Cobbe et~al.(2021)Cobbe, Kosaraju, Bavarian, Chen, Jun, Kaiser, Plappert, Tworek, Hilton, Nakano et~al.}]{openai_math}
Karl Cobbe, Vineet Kosaraju, Mohammad Bavarian, Mark Chen, Heewoo Jun, Lukasz Kaiser, Matthias Plappert, Jerry Tworek, Jacob Hilton, Reiichiro Nakano, et~al. 2021.
\newblock Training verifiers to solve math word problems.
\newblock \emph{arXiv preprint arXiv:2110.14168}.

\bibitem[{Dijksterhuis(2005)}]{dijksterhuis2005we}
Ap~Dijksterhuis. 2005.
\newblock Why we are social animals: The high road to imitation as social glue.
\newblock \emph{Perspectives on imitation: From neuroscience to social science}, 2:207--220.

\bibitem[{El-Kassas et~al.(2021)El-Kassas, Salama, Rafea, and Mohamed}]{el2021automatic}
Wafaa~S El-Kassas, Cherif~R Salama, Ahmed~A Rafea, and Hoda~K Mohamed. 2021.
\newblock Automatic text summarization: A comprehensive survey.
\newblock \emph{Expert systems with applications}, 165:113679.

\bibitem[{Fan et~al.(2021)Fan, Bhosale, Schwenk, Ma, El-Kishky, Goyal, Baines, Celebi, Wenzek, Chaudhary et~al.}]{fan2021beyond}
Angela Fan, Shruti Bhosale, Holger Schwenk, Zhiyi Ma, Ahmed El-Kishky, Siddharth Goyal, Mandeep Baines, Onur Celebi, Guillaume Wenzek, Vishrav Chaudhary, et~al. 2021.
\newblock Beyond english-centric multilingual machine translation.
\newblock \emph{Journal of Machine Learning Research}, 22(107):1--48.

\bibitem[{Goody(1995)}]{goody1995social}
Esther~N Goody. 1995.
\newblock \emph{Social intelligence and interaction: Expressions and implications of the social bias in human intelligence}.
\newblock Cambridge University Press.

\bibitem[{Gygax and Arneson(1974)}]{gygax1974dungeons}
Gary Gygax and Dave Arneson. 1974.
\newblock \emph{dungeons \& dragons}, volume~19.
\newblock Tactical Studies Rules Lake Geneva, WI.

\bibitem[{Heikkinen et~al.(2021)Heikkinen, L{\"a}ms{\"a}, and Niemist{\"o}}]{heikkinen2021work}
Suvi Heikkinen, Anna-Maija L{\"a}ms{\"a}, and Charlotta Niemist{\"o}. 2021.
\newblock Work--family practices and complexity of their usage: a discourse analysis towards socially responsible human resource management.
\newblock \emph{Journal of Business Ethics}, 171:815--831.

\bibitem[{Hendrycks et~al.(2020)Hendrycks, Burns, Basart, Zou, Mazeika, Song, and Steinhardt}]{MMLU}
Dan Hendrycks, Collin Burns, Steven Basart, Andy Zou, Mantas Mazeika, Dawn Song, and Jacob Steinhardt. 2020.
\newblock Measuring massive multitask language understanding.
\newblock \emph{arXiv preprint arXiv:2009.03300}.

\bibitem[{Huang and Chang(2022)}]{huang2022towards}
Jie Huang and Kevin Chen-Chuan Chang. 2022.
\newblock Towards reasoning in large language models: A survey.
\newblock \emph{arXiv preprint arXiv:2212.10403}.

\bibitem[{Huang et~al.(2023)Huang, Bai, Zhu, Zhang, Zhang, Su, Liu, Lv, Zhang, Lei et~al.}]{ceval}
Yuzhen Huang, Yuzhuo Bai, Zhihao Zhu, Junlei Zhang, Jinghan Zhang, Tangjun Su, Junteng Liu, Chuancheng Lv, Yikai Zhang, Jiayi Lei, et~al. 2023.
\newblock C-eval: A multi-level multi-discipline chinese evaluation suite for foundation models.
\newblock \emph{arXiv preprint arXiv:2305.08322}.

\bibitem[{Imani et~al.(2023)Imani, Du, and Shrivastava}]{imani2023mathprompter}
Shima Imani, Liang Du, and Harsh Shrivastava. 2023.
\newblock Mathprompter: Mathematical reasoning using large language models.
\newblock \emph{arXiv preprint arXiv:2303.05398}.

\bibitem[{Lehnert(2022)}]{lehnert2022process}
Wendy~G Lehnert. 2022.
\newblock \emph{The process of question answering: A computer simulation of cognition}.
\newblock Taylor \& Francis.

\bibitem[{Li et~al.(2023)Li, Song, Yu, Yu, Li, Huang, and Li}]{apibank}
Minghao Li, Feifan Song, Bowen Yu, Haiyang Yu, Zhoujun Li, Fei Huang, and Yongbin Li. 2023.
\newblock Api-bank: A benchmark for tool-augmented llms.
\newblock \emph{arXiv preprint arXiv:2304.08244}.

\bibitem[{Liang et~al.(2022)Liang, Bommasani, Lee, Tsipras, Soylu, Yasunaga, Zhang, Narayanan, Wu, Kumar et~al.}]{HELM}
Percy Liang, Rishi Bommasani, Tony Lee, Dimitris Tsipras, Dilara Soylu, Michihiro Yasunaga, Yian Zhang, Deepak Narayanan, Yuhuai Wu, Ananya Kumar, et~al. 2022.
\newblock Holistic evaluation of language models.
\newblock \emph{arXiv preprint arXiv:2211.09110}.

\bibitem[{Liang et~al.(2023)Liang, Zhu, and Yang}]{liang2023tachikuma}
Yuanzhi Liang, Linchao Zhu, and Yi~Yang. 2023.
\newblock Tachikuma: Understading complex interactions with multi-character and novel objects by large language models.
\newblock \emph{arXiv preprint arXiv:2307.12573}.

\bibitem[{Lipenkova(2023)}]{lipenkova2023overcoming}
Janna Lipenkova. 2023.
\newblock Overcoming the limitations of large language models how to enhance llms with human-like cognitive skills.

\bibitem[{Liu et~al.(2023{\natexlab{a}})Liu, Yuan, Fu, Jiang, Hayashi, and Neubig}]{llm_survey}
Pengfei Liu, Weizhe Yuan, Jinlan Fu, Zhengbao Jiang, Hiroaki Hayashi, and Graham Neubig. 2023{\natexlab{a}}.
\newblock Pre-train, prompt, and predict: A systematic survey of prompting methods in natural language processing.
\newblock \emph{ACM Computing Surveys}, 55(9):1--35.

\bibitem[{Liu et~al.(2023{\natexlab{b}})Liu, Yu, Zhang, Xu, Lei, Lai, Gu, Ding, Men, Yang et~al.}]{agentbench}
Xiao Liu, Hao Yu, Hanchen Zhang, Yifan Xu, Xuanyu Lei, Hanyu Lai, Yu~Gu, Hangliang Ding, Kaiwen Men, Kejuan Yang, et~al. 2023{\natexlab{b}}.
\newblock Agentbench: Evaluating llms as agents.
\newblock \emph{arXiv preprint arXiv:2308.03688}.

\bibitem[{Lopes et~al.(2004)Lopes, Brackett, Nezlek, Sch{\"u}tz, Sellin, and Salovey}]{lopes2004emotional}
Paulo~N Lopes, Marc~A Brackett, John~B Nezlek, Astrid Sch{\"u}tz, Ina Sellin, and Peter Salovey. 2004.
\newblock Emotional intelligence and social interaction.
\newblock \emph{Personality and social psychology bulletin}, 30(8):1018--1034.

\bibitem[{Louis and Sutton(2018)}]{louis2018deep}
Annie Louis and Charles Sutton. 2018.
\newblock Deep dungeons and dragons: Learning character-action interactions from role-playing game transcripts.
\newblock In \emph{Proceedings of the 2018 Conference of the North American Chapter of the Association for Computational Linguistics: Human Language Technologies, Volume 2 (Short Papers)}, pages 708--713.

\bibitem[{Lu et~al.(2022)Lu, Clark, Zellers, Mottaghi, and Kembhavi}]{lu2022unified}
Jiasen Lu, Christopher Clark, Rowan Zellers, Roozbeh Mottaghi, and Aniruddha Kembhavi. 2022.
\newblock Unified-io: A unified model for vision, language, and multi-modal tasks.
\newblock \emph{arXiv preprint arXiv:2206.08916}.

\bibitem[{Newman and Liu(2022)}]{newman-liu-2022-generating}
Pax Newman and Yudong Liu. 2022.
\newblock \href {https://aclanthology.org/2022.games-1.7} {Generating descriptive and rules-adhering spells for dungeons {\&} dragons fifth edition}.
\newblock In \emph{Proceedings of the 9th Workshop on Games and Natural Language Processing within the 13th Language Resources and Evaluation Conference}, pages 54--60, Marseille, France. European Language Resources Association.

\bibitem[{Park et~al.(2023)Park, O'Brien, Cai, Morris, Liang, and Bernstein}]{park2023generative}
Joon~Sung Park, Joseph~C O'Brien, Carrie~J Cai, Meredith~Ringel Morris, Percy Liang, and Michael~S Bernstein. 2023.
\newblock Generative agents: Interactive simulacra of human behavior.
\newblock \emph{arXiv preprint arXiv:2304.03442}.

\bibitem[{Rawte et~al.(2023)Rawte, Sheth, and Das}]{rawte2023survey}
Vipula Rawte, Amit Sheth, and Amitava Das. 2023.
\newblock A survey of hallucination in large foundation models.
\newblock \emph{arXiv preprint arXiv:2309.05922}.

\bibitem[{Sakaguchi et~al.(2021)Sakaguchi, Bras, Bhagavatula, and Choi}]{winogrande}
Keisuke Sakaguchi, Ronan~Le Bras, Chandra Bhagavatula, and Yejin Choi. 2021.
\newblock Winogrande: An adversarial winograd schema challenge at scale.
\newblock \emph{Communications of the ACM}, 64(9):99--106.

\bibitem[{Si et~al.(2021)Si, Ammanabrolu, and Riedl}]{si-etal-2021-telling}
Wai~Man Si, Prithviraj Ammanabrolu, and Mark Riedl. 2021.
\newblock \href {https://aclanthology.org/2021.sigdial-1.30} {Telling stories through multi-user dialogue by modeling character relations}.
\newblock In \emph{Proceedings of the 22nd Annual Meeting of the Special Interest Group on Discourse and Dialogue}, pages 269--275, Singapore and Online. Association for Computational Linguistics.

\bibitem[{Srivastava et~al.(2022)Srivastava, Rastogi, Rao, Shoeb, Abid, Fisch, Brown, Santoro, Gupta, Garriga-Alonso et~al.}]{bigbench}
Aarohi Srivastava, Abhinav Rastogi, Abhishek Rao, Abu Awal~Md Shoeb, Abubakar Abid, Adam Fisch, Adam~R Brown, Adam Santoro, Aditya Gupta, Adri{\`a} Garriga-Alonso, et~al. 2022.
\newblock Beyond the imitation game: Quantifying and extrapolating the capabilities of language models.
\newblock \emph{arXiv preprint arXiv:2206.04615}.

\bibitem[{Sterelny(2007)}]{sterelny2007social}
Kim Sterelny. 2007.
\newblock Social intelligence, human intelligence and niche construction.
\newblock \emph{Philosophical Transactions of the Royal Society B: Biological Sciences}, 362(1480):719--730.

\bibitem[{Valmeekam et~al.(2022)Valmeekam, Olmo, Sreedharan, and Kambhampati}]{valmeekam2022large}
Karthik Valmeekam, Alberto Olmo, Sarath Sreedharan, and Subbarao Kambhampati. 2022.
\newblock Large language models still can't plan (a benchmark for llms on planning and reasoning about change).
\newblock \emph{arXiv preprint arXiv:2206.10498}.

\bibitem[{Wang et~al.(2023)Wang, Zhou, Xu, Shi, Zhao, Xu, Ye, Yan, Zhang, Zhu et~al.}]{wang2023evaluation}
Junyang Wang, Yiyang Zhou, Guohai Xu, Pengcheng Shi, Chenlin Zhao, Haiyang Xu, Qinghao Ye, Ming Yan, Ji~Zhang, Jihua Zhu, et~al. 2023.
\newblock Evaluation and analysis of hallucination in large vision-language models.
\newblock \emph{arXiv preprint arXiv:2308.15126}.

\bibitem[{Weir et~al.(2022)Weir, Thomas, D'Amore, Hill, Van~Durme, and Jhamtani}]{weir2022ontologically}
Nathaniel Weir, Ryan Thomas, Randolph D'Amore, Kellie Hill, Benjamin Van~Durme, and Harsh Jhamtani. 2022.
\newblock Ontologically faithful generation of non-player character dialogues.
\newblock \emph{arXiv preprint arXiv:2212.10618}.

\bibitem[{Widyassari et~al.(2022)Widyassari, Rustad, Shidik, Noersasongko, Syukur, Affandy et~al.}]{widyassari2022review}
Adhika~Pramita Widyassari, Supriadi Rustad, Guruh~Fajar Shidik, Edi Noersasongko, Abdul Syukur, Affandy Affandy, et~al. 2022.
\newblock Review of automatic text summarization techniques \& methods.
\newblock \emph{Journal of King Saud University-Computer and Information Sciences}, 34(4):1029--1046.

\bibitem[{Xu et~al.(2023)Xu, Li, Zhu, Xue, Zhu, Zhao, He, Zhang, Kang, and Lan}]{superclue}
Liang Xu, Anqi Li, Lei Zhu, Hang Xue, Changtai Zhu, Kangkang Zhao, Haonan He, Xuanwei Zhang, Qiyue Kang, and Zhenzhong Lan. 2023.
\newblock Superclue: A comprehensive chinese large language model benchmark.
\newblock \emph{arXiv preprint arXiv:2307.15020}.

\bibitem[{Yang et~al.(2020)Yang, Wang, and Chu}]{yang2020survey}
Shuoheng Yang, Yuxin Wang, and Xiaowen Chu. 2020.
\newblock A survey of deep learning techniques for neural machine translation.
\newblock \emph{arXiv preprint arXiv:2002.07526}.

\bibitem[{Ye et~al.(2023)Ye, Chen, Dillig, and Durrett}]{ye2023satlm}
Xi~Ye, Qiaochu Chen, Isil Dillig, and Greg Durrett. 2023.
\newblock Satlm: Satisfiability-aided language models using declarative prompting.
\newblock \emph{Proceedings of NeurIPS}, pages 1--33.

\bibitem[{Zellers et~al.(2019)Zellers, Holtzman, Bisk, Farhadi, and Choi}]{hellaswag}
Rowan Zellers, Ari Holtzman, Yonatan Bisk, Ali Farhadi, and Yejin Choi. 2019.
\newblock Hellaswag: Can a machine really finish your sentence?
\newblock \emph{arXiv preprint arXiv:1905.07830}.

\bibitem[{Zhong et~al.(2023)Zhong, Cui, Guo, Liang, Lu, Wang, Saied, Chen, and Duan}]{agieval}
Wanjun Zhong, Ruixiang Cui, Yiduo Guo, Yaobo Liang, Shuai Lu, Yanlin Wang, Amin Saied, Weizhu Chen, and Nan Duan. 2023.
\newblock Agieval: A human-centric benchmark for evaluating foundation models.
\newblock \emph{arXiv preprint arXiv:2304.06364}.

\bibitem[{Zhou et~al.(2022)Zhou, Zhu, Hu, Pujara, Ren, Callison-Burch, Choi, and Ammanabrolu}]{gandalf}
Pei Zhou, Andrew Zhu, Jennifer Hu, Jay Pujara, Xiang Ren, Chris Callison-Burch, Yejin Choi, and Prithviraj Ammanabrolu. 2022.
\newblock An ai dungeon master's guide: Learning to converse and guide with intents and theory-of-mind in dungeons and dragons.
\newblock \emph{arXiv preprint arXiv:2212.10060}.

\bibitem[{Zhu et~al.(2023)Zhu, Aggarwal, Feng, Martin, and Callison-Burch}]{zhu2023fireball}
Andrew Zhu, Karmanya Aggarwal, Alexander Feng, Lara~J Martin, and Chris Callison-Burch. 2023.
\newblock Fireball: A dataset of dungeons and dragons actual-play with structured game state information.
\newblock \emph{arXiv preprint arXiv:2305.01528}.

\bibitem[{Zhu et~al.(2021)Zhu, Lei, Wang, Zheng, Poria, and Chua}]{zhu2021retrieving}
Fengbin Zhu, Wenqiang Lei, Chao Wang, Jianming Zheng, Soujanya Poria, and Tat-Seng Chua. 2021.
\newblock Retrieving and reading: A comprehensive survey on open-domain question answering.
\newblock \emph{arXiv preprint arXiv:2101.00774}.

\end{thebibliography}
